\documentclass[runningheads]{llncs}
\usepackage{graphicx}
\usepackage{comment}
\usepackage{caption}
\usepackage{subcaption}
\usepackage{dingbat} 
\begin{document}
\title{Lyric Video Analysis \\
Using Text Detection and Tracking}
\author{Shota Sakaguchi\inst{1}
\and
Jun Kato\inst{2}
\and
Masataka Goto\inst{2}
\and
Seiichi Uchida\inst{1}
}
\authorrunning{Sakaguchi et al.}
\institute{Kyushu University, Fukuoka, Japan
\and
National Institute of Advanced Industrial Science and Technology (AIST)
\email{
\{shota.sakaguchi, uchida\}@human.ait.kyushu-u.ac.jp
}\\
\email{
\{jun.kato, m.goto\}@aist.go.jp
}
}

\maketitle 
\begin{abstract}
We attempt to recognize and track lyric words in lyric videos. Lyric video is a music video showing the lyric words of a song. The main characteristic of lyric videos is that the lyric words are shown at frames synchronously with the music. The difficulty of recognizing and tracking the lyric words is that (1) the words are often decorated and geometrically distorted and (2) the words move arbitrarily and drastically in the video frame. The purpose of this paper is to analyze the motion of the lyric words in lyric videos, as the first step of automatic lyric video generation. In order to analyze the motion of lyric words, we first apply a state-of-the-art scene text detector and recognizer to each video frame. Then, lyric-frame matching is performed to establish the optimal correspondence between lyric words and the frames. After fixing the motion trajectories of individual lyric words from correspondence, we analyze the trajectories of the lyric words by k-medoids clustering and dynamic time warping (DTW).

\keywords{Lyric video \and Lyric word tracking \and Text motion analysis \and Video design analysis.}
\end{abstract}

\section{Introduction\label{sec:intro}}
The targets of document analysis systems are expanding because of the diversity of
recent document modalities.
Scanned paper documents only with texts printed in ordinary fonts were
traditional targets of the systems.  However, recent advanced camera-based OCR
technologies allow us to analyze arbitrary images and extract text information
from them. In other words, we can now treat arbitrary images with text information as documents.
\par
In fact, we can consider videos as a promising target of document analysis systems.
There have already been many attempts to analyze videos as
documents~\cite{yin2016text}.
The most typical attempt is caption detection and recognition in video frames. Another attempt is the analysis of the video from the in-vehicle camera. By recognizing the texts in the in-vehicle videos, it is possible to collect store and building names and signboard information around the road automatically.
There was also an attempt (e.g., \cite{soccer} as a classical trial) to recognize the sport player's jersey number for identifying individual players and then analyzing their performance.
\par
\begin{figure}[t]
  \begin{minipage}[t]{0.5\linewidth}
    \centering
    \includegraphics[width=0.8\textwidth]{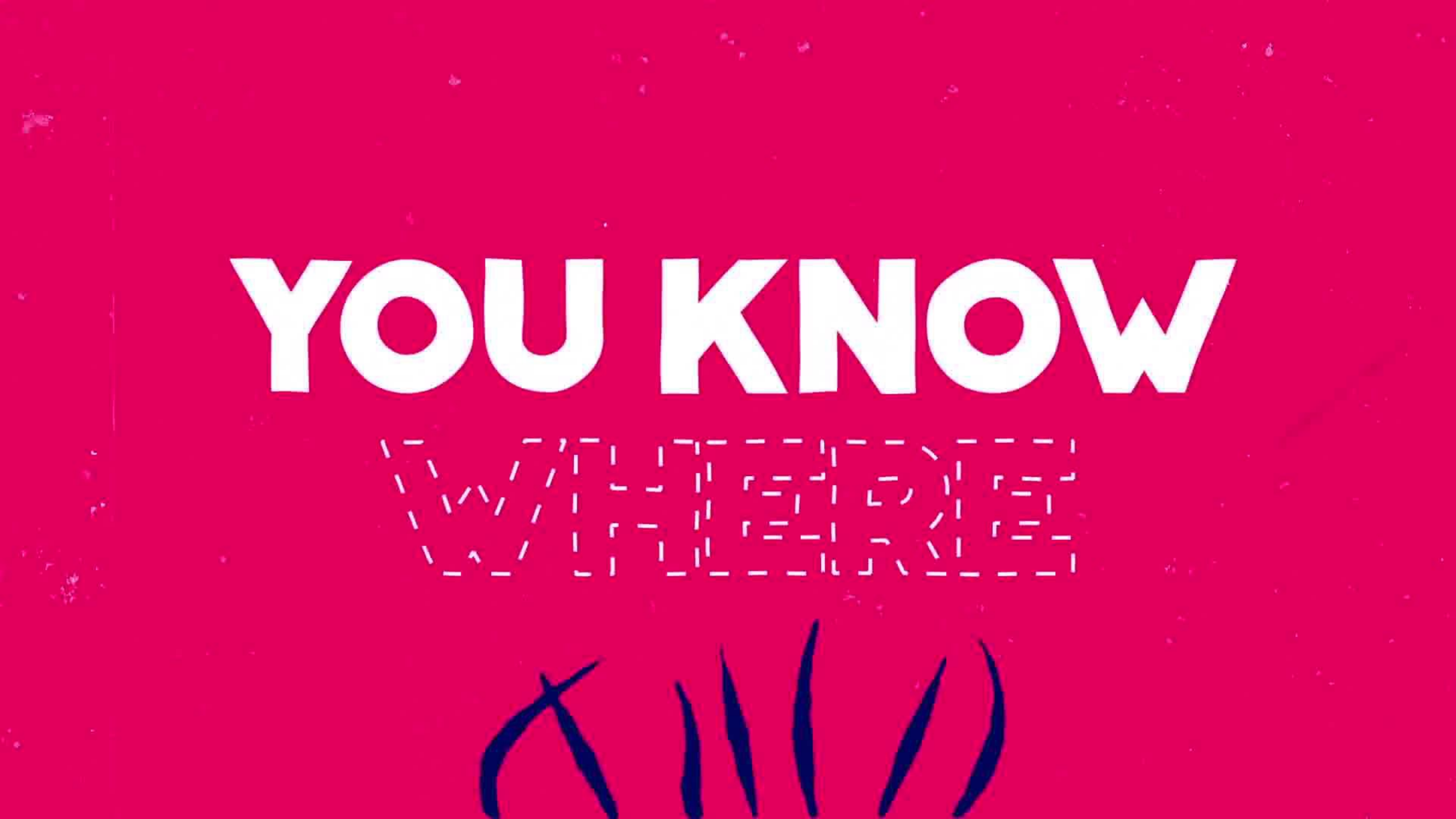}\\[-2mm]
    \subcaption{``YOU KNOW WHERE''}
    \label{lyric_video_examples1}
  \end{minipage}
  \begin{minipage}[t]{0.5\linewidth}
    \centering
    \includegraphics[width=0.8\textwidth]{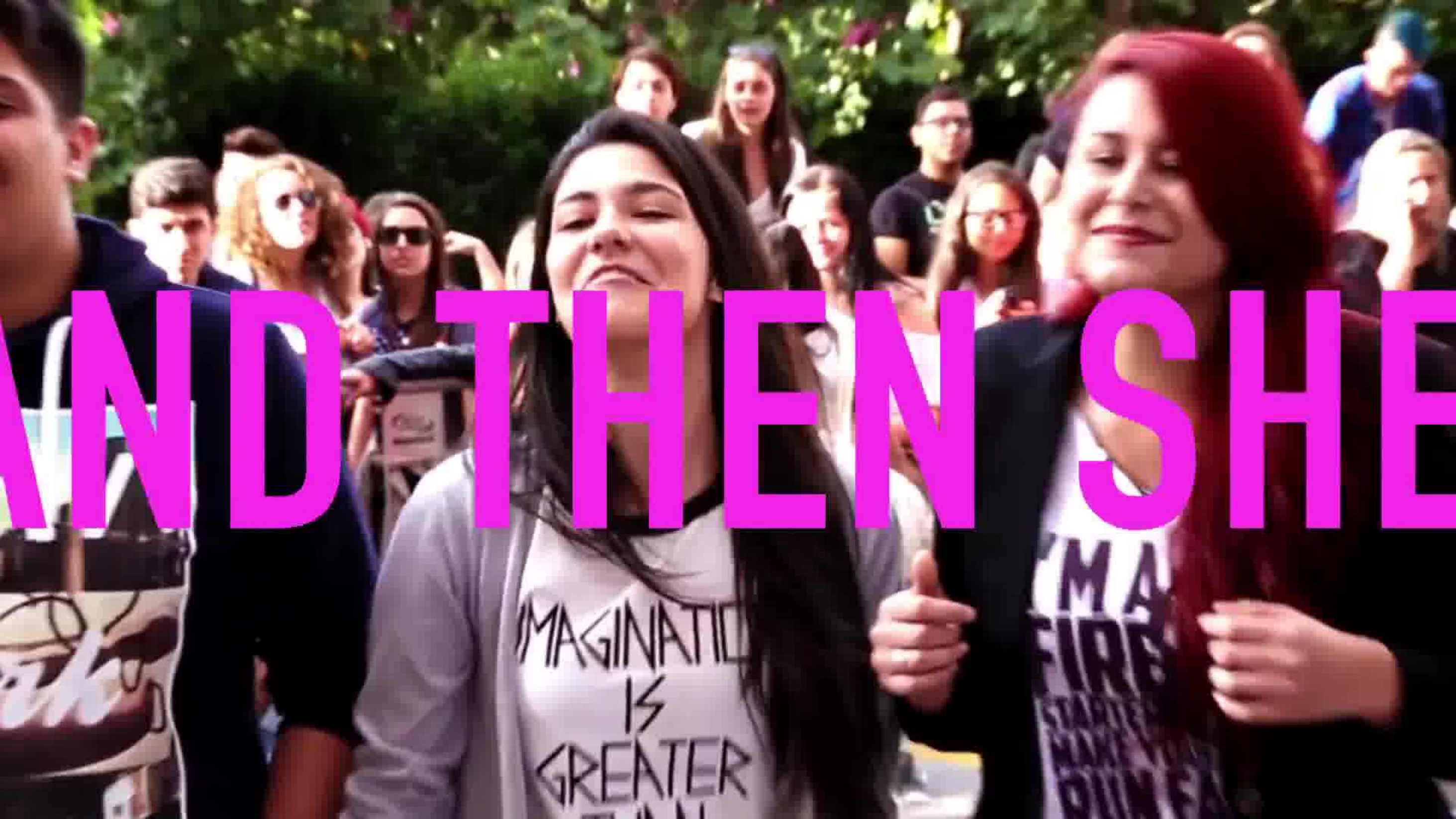}\\[-2mm]
    \subcaption{``AND THEN SHE''}
    \label{lyric_video_examples2}
  \end{minipage}\par
 \bigskip
  \begin{minipage}[t]{0.5\linewidth}
    \centering
    \includegraphics[width=0.8\textwidth]{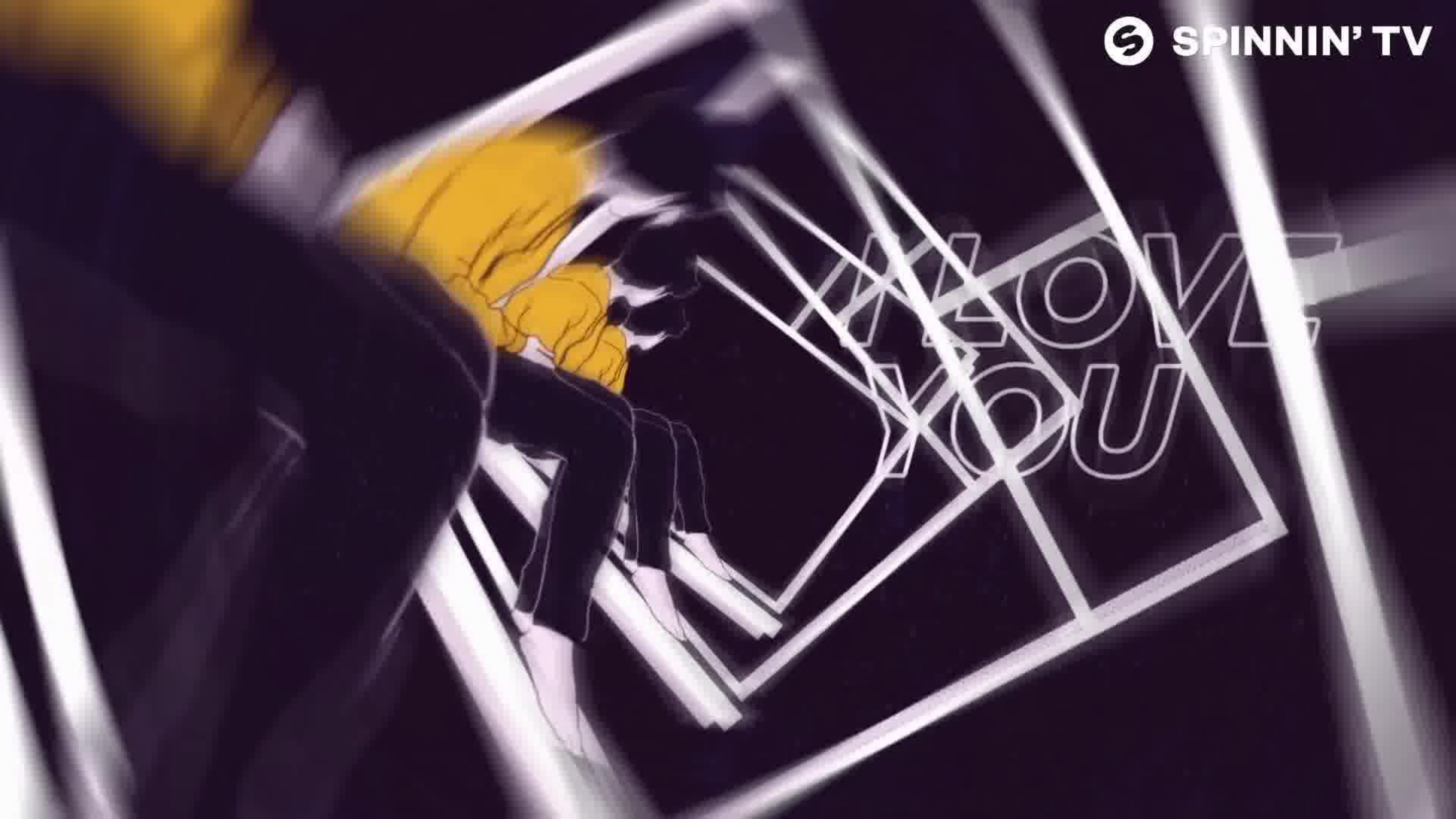}\\[-2mm]
    \subcaption{``I LOVE YOU''}
    \label{lyric_video_examples3}
  \end{minipage}
  \begin{minipage}[t]{0.5\linewidth}
    \centering
    \includegraphics[width=0.8\textwidth]{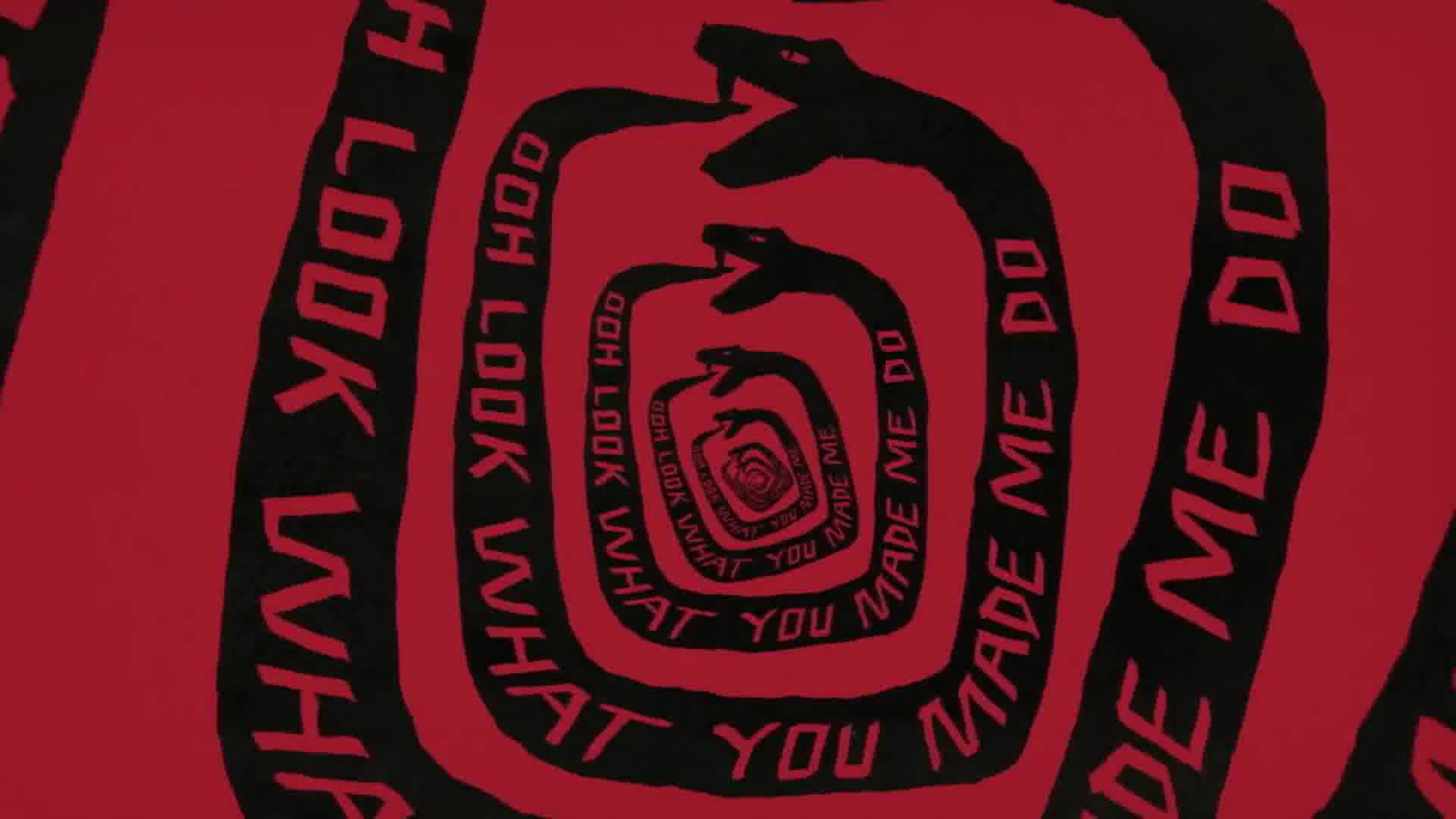}\\[-2mm]
    \subcaption{``OOH LOOK WHAT YOU MADE ME DO''}
    \label{lyric_video_examples4}
  \end{minipage}
  \caption{Frame image examples from lyric video.}
  \label{fig:lyric_video_examples}
\end{figure}

In this paper, we use {\em lyric videos} as a new target of document
analysis. Lyric videos are music videos published at internet video services,
such as {\tt YouTube}, for promoting a song. The main characteristic of lyric
videos is that they show the lyric words of the song (almost) synchronously to
the music. Fig.~\ref{fig:lyric_video_examples} shows several frame examples from
lyric videos. Lyric words are often printed in various decorated fonts and
distorted by elaborated visual designs; they, therefore, are
sometimes hard to read even for humans. Background images of the frames are
photographic images or illustrations or their mixtures and often too complex
to read the lyrics. This means that lyric word detection and recognition for
lyric videos is a difficult task even for state-of-the-art scene text detectors
and recognizers.\par

In addition, lyric words in lyric videos move very differently from words in ordinary videos.
For example, lyric words often move along with arbitrarily-shaped trajectories while rotating and scaling. This property
is very different from video captions since they do not move or just scroll in video frames.
It is also different from scene texts in videos from the in-vehicle camera;
scene texts move passively according to camera motion, whereas lyric words move
actively and arbitrarily.
In fact, lyric words and their motion are designed carefully by the creator of the lyric video
so that the motion makes the video more impressive. To give a strong impression,
lyric words might move like an explosion. \par

The purpose of this paper is to analyze the motion of the lyric words in lyric
videos. Specifically, we try to extract and track words in the lyric words automatically
and then group (i.e., clustering) their motions into several types to understand their trends.
Nowadays, anyone can post her/his music with video. This research
will contribute to those non-professional creators to design their own lyric
videos. In the future, it will also be possible to realize automatic lyric video
generation, by extending this work to understand the relationship between lyric
word motion and music beat or music style.
\par
\begin{figure}[t]
	\begin{center}
	\includegraphics[width=1.0\textwidth]{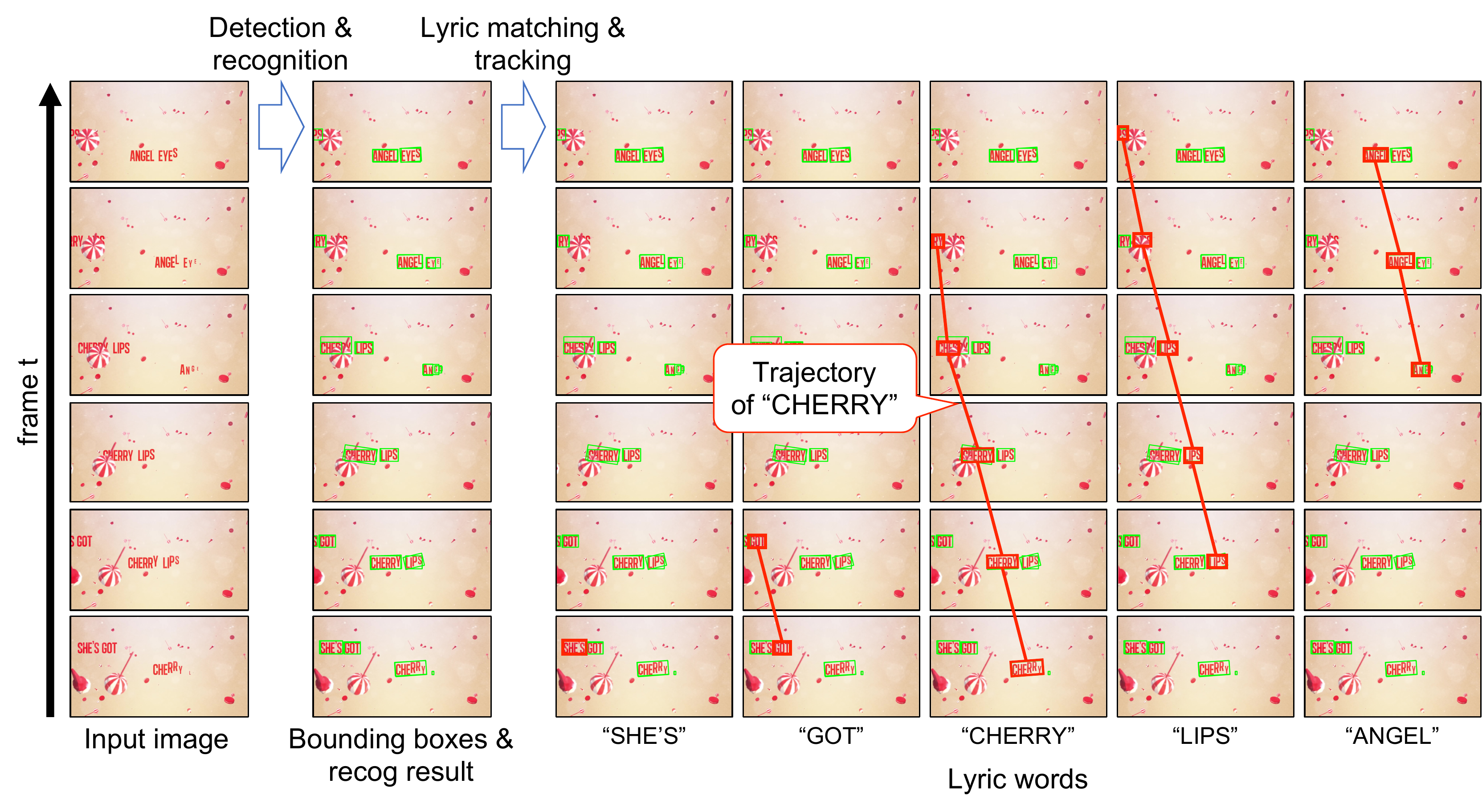}
	\caption{Our lyric video analysis task. After detecting and recognizing
	 words in video frames, the words in the lyric are tracked over
	 frames. Finally, the word motion trajectories are classified into
	 several clusters to understand the trends of lyric word motions.}
		\label{fig:overall}
	\end{center}
\end{figure}

For the above purpose, we develop a lyric word detection and tracking method,
where lyric information is fully utilized. As noted before, lyric words are heavily decorated and
shown on various background images. Even state-of-the-art scene text detectors
and recognizers will have miss-detections (false-negatives), false-detections
(false-positives), and erroneous word recognition results.  We, therefore, utilize the lyric
information as metadata for improving detection and tracking performance. As
shown in Fig.~\ref{fig:overall}, we first apply state-of-the-art scene text
detectors and recognizers to each video frame. Then, dynamic programming-based
optimization is applied to determine the optimal matching between the video frame
sequence and the lyric word sequence. We call this novel technique
{\em lyric-frame matching}. Then, using the lyric-frame matching result
as a reliable correspondence, spatio-temporal search (tracking and interpolation) is performed
for fixing the motion trajectory of each lyric word.

\par

The main contributions of this paper are summarized as follows:
\begin{itemize}
    \item First, to the authors' best knowledge, this is the first trial of
	  detecting and then tracking lyric words in lyric videos.
    \item Second, we propose a lyric-frame matching technique
	  where lyric information of the target music is fully utilized as
	  metadata for accurately determining the most confident frame
	  where a lyric word appears. Dynamic programming and state-of-the-art text detection
	  and recognition techniques are used as the modules of the proposed technique.
    \item Third, we prepared a dataset comprised of 100 lyric videos and
	  attached ground-truth (bounding boxes for lyric words) manually at
	  10 frames for each of 100 videos. This effort enables
	  us to make a quantitative evaluation of our detection and tracking
	  results.
    \item Fourth, as an analysis of lyric word motions, we grouped them into
	  several representative word motions by k-medoid clustering. From the clustering result, we could
	  represent each lyric video by a histogram of representative motions.
\end{itemize}
\par

\section{Related Work\label{sec:related}}
To the authors' best knowledge, this paper is the first trial of analyzing lyric videos as a new document modality. We, however, still can find several related tasks, that is, video caption detection and recognition, and text tracking. In the following, we review the past attempts on those tasks, although the readers also refer to a comprehensive survey~\cite{yin2016text,ijdar}. It should be noted that the performance on all those tasks has drastically been improved by the recent progress of scene text detection and recognition technologies. Since the attempts on scene text detection and recognition are so huge (even if we limit them only to the recent ones), we will not make any review on the topic. A survey~\cite{ijdar} will give its good overview.
\par
Caption detection and recognition is a classical but still hot topic of document image analysis research. Captions are defined as the texts superimposed on video frames. Captions, therefore, have different characteristics from scene texts.
We can find many attempts on this task, such as \cite{zhao2010text,zhong2000automatic,yang2012caption,lu2014video,yang2014framework,zhong2016recognition,zedan2016caption,chen2018video,xu2018end,lu2018novel}.
Most of them deal with the static captions (i.e., captions without motions), while Zedan~\cite{zedan2016caption} deals with moving captions; they assume the vertical or horizontal scrolling of caption text, in addition to the static captions.
\par
Rather recently, scene text tracking and video text tracking [14,15,16,17,18,19,\\20,21,22] have also been tried, as reviewed in \cite{yin2016text}. Each method introduces its own word tracker. For example,  Yang~\cite{yang2017tracking} uses a dynamic programming-based tracker to have an optimal spatio-temporal trajectory for each word. A common assumption underlying those attempts is that text motions in video frames are caused mainly by camera motion. Therefore, for example, neighboring words will move to similar directions.   We also find ``moving MNIST''\cite{movingmnist} for a video prediction task, but it just deals with synthetic videos capturing two digits are moving around on a uniform background.

\par
Our trial is very different from those past attempts at the following three points at least. First,
our target texts move far more dynamically than the texts that the past attempts assume. In fact, we cannot make any assumption on the location, appearance, and motion of the words.
Second, our trial can utilize the lyric information as the reliable guide of text tracking and therefore we newly propose a lyric-frame matching technique for our task. Third, our main focus is to analyze the text motion trajectory from a viewpoint of video design analysis, which is totally ignored in the past text tracking attempts.\par

\section{Lyric Video Dataset\label{sec:dataset}}
We collected 100 lyric videos according to the following steps.
First, a list of lyric videos is generated by searching YouTube with the keyword
``official lyric video''~\footnote{The search was performed on 18th July
2019.}. The keyword ``official'' was added not only for finding videos with
long-term availability and but also for finding videos created by
professional creators. Videos on the list were then checked visually. A video only
with static lyric words (i.e., a video whose lyric words do not move) is removed from the
list. Finally, the top-100 videos on the list are selected as our targets\footnote{For
the URL list of all the videos and their annotation data, please refer to
{\tt https://github.com/uchidalab/Lyric-Video}.}.
\par

The collected 100 lyric videos have the following statistics. Each video is
comprised of 5,471 frames (3 min 38 sec) on average, 8,629 frames (5 min 44 sec)
at maximum, and 2,280 frames (2 min 33 sec) at minimum. The frame image size is
1,920 $\times$ 1,080 pixels. The frame rate is 12, 24, 25, and 30 fps
on 1, 58, 25, and 16 videos, respectively. The lyrics are basically in
English. Each song is comprised of 338 lyric words on average, 690 words at
maximum, and 113 at minimum.
\par

\begin{figure}[t]
  \begin{minipage}[t]{0.5\linewidth}
    \centering
    \includegraphics[width=0.8\textwidth]{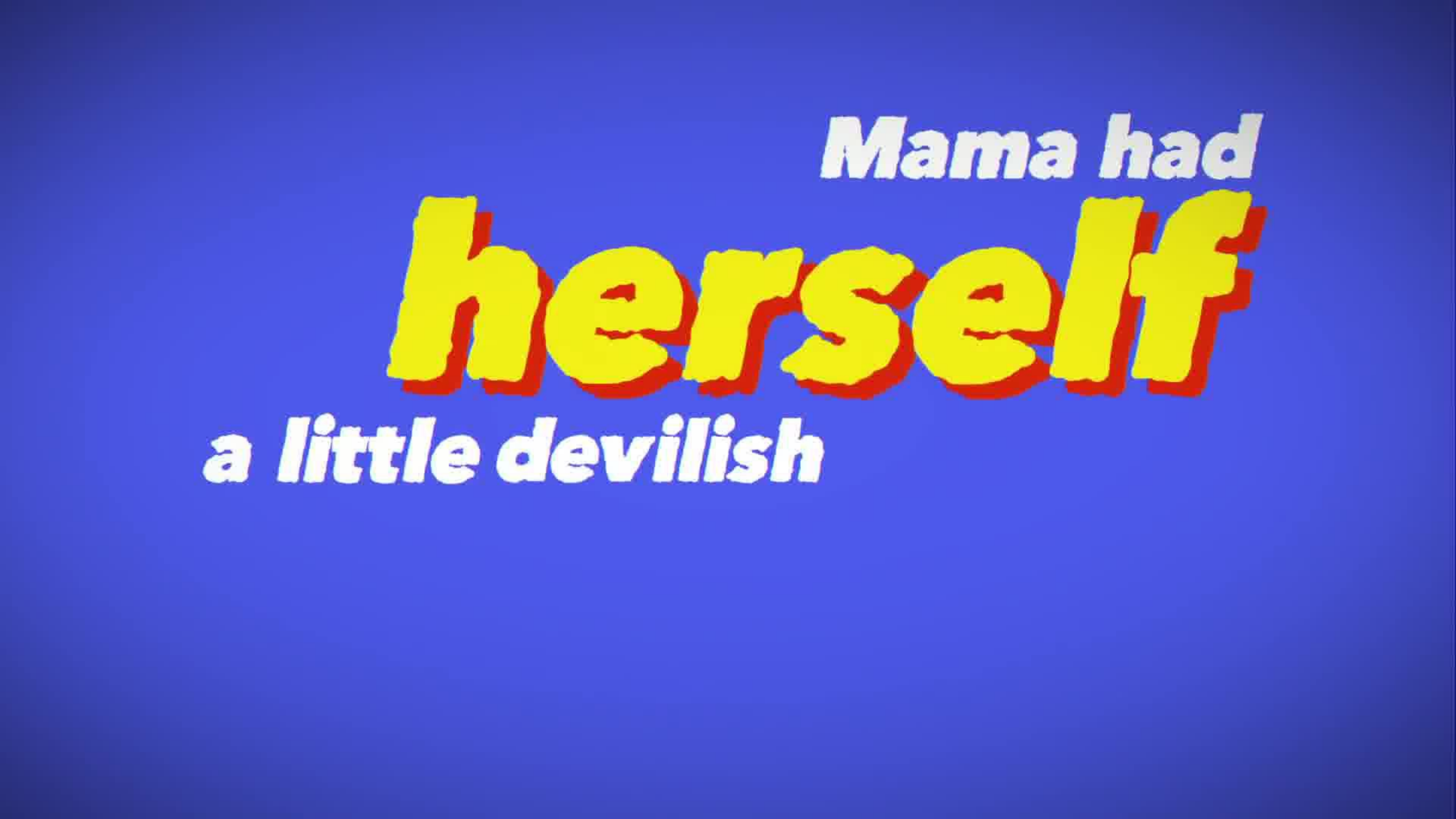}\\[-2mm]
    \subcaption{Showing lyric words.}
  \end{minipage}
  \begin{minipage}[t]{0.5\linewidth}
    \centering
    \includegraphics[width=0.8\textwidth]{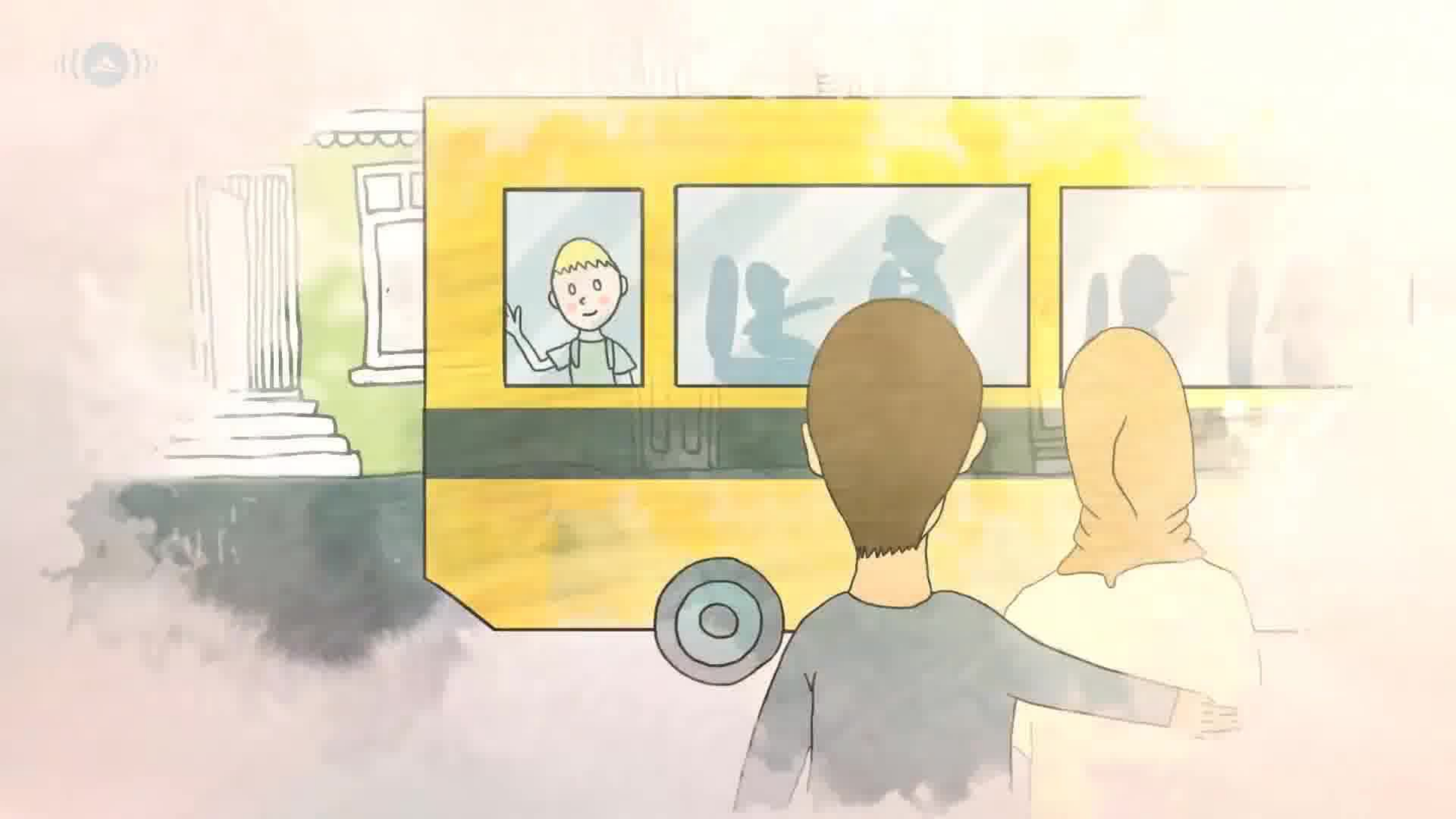}\\[-2mm]
    \subcaption{No lyric word (interlude).}
  \end{minipage}\par
  \bigskip
  \begin{minipage}[t]{0.5\linewidth}
    \centering
    \includegraphics[width=0.8\textwidth]{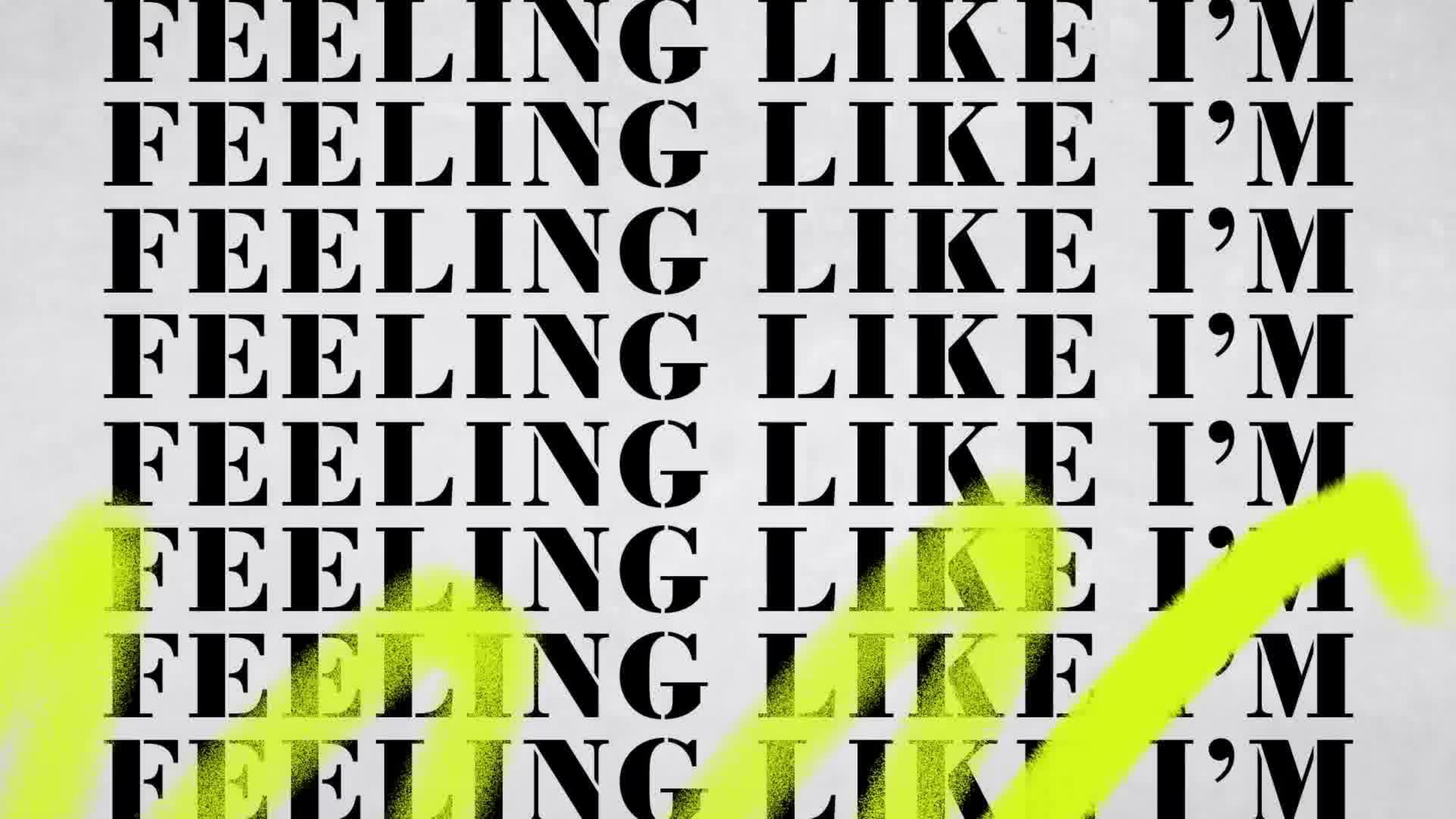}\\[-2mm]
    \subcaption{Duplicated lyric words (like refrains).}
  \end{minipage}
  \begin{minipage}[t]{0.5\linewidth}
    \centering
    \includegraphics[width=0.8\textwidth]{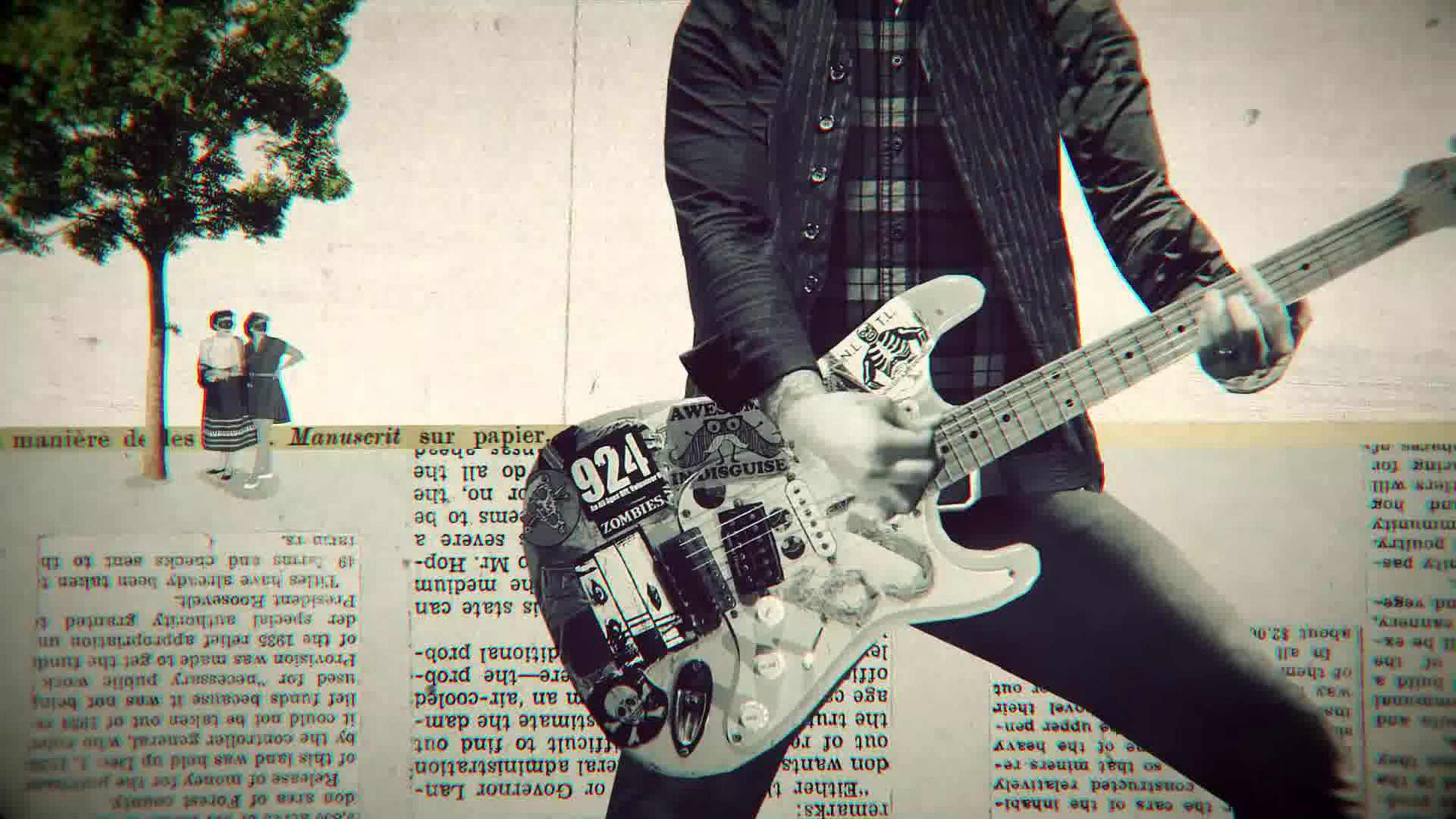}\\[-2mm]
    \subcaption{Words unrelated to the lyrics.}
  \end{minipage}
  \caption{Variations of the spatial distribution of lyric words in a video frame.}
  \label{fig:lyric_words_example}
\end{figure}

Fig.~\ref{fig:lyric_words_example} shows four examples showing the variations
of the number of lyric words in a frame. Fig.~\ref{fig:lyric_words_example}~(a) shows a
typical frame with lyric words. It is quite often in lyric videos that a phrase
(i.e., consecutive lyric words) is shown over several frames. It is rather rare
that only a single lyric word is shown in a one-by-one manner synchronously with
the music. We also have frames without any lyric word, like (b). Those frames are often found in
introduction parts, interlude parts, and ending parts of lyric videos.
Figs.~\ref{fig:lyric_words_example}~(c) and (d) contain many words in a frame. In
(c), the same word is repeatedly shown in a frame like a refrain of a song. The
words in (d) are just decorations (i.e., they belong to the background)
and not relating to the lyrics.  \par
For a quantitative evaluation in our experiment, we attach the bounding boxes for
lyric words manually at 10 frames for each lyric video. The 10 frames were
selected as follows; for each video, we picked up frames every three seconds and
then select 10 frames with the most word candidates detected automatically
by the procedure of Section~\ref{sec:detection}. The three-second interval is
necessary to avoid the selection of the consecutive frames.
If a lyric word appears multiple times in a frame (like
Fig.~\ref{fig:lyric_words_example}~(c)), the bounding box is attached to each of
them. For the rotated lyric words, we attached a non-horizontal bounding
box~\footnote{To attach non-horizontal bounding boxes,
we used the labeling tool {\tt roLabelImg} at {\tt https://github.com/cgvict/roLabelImg.}}.
Consequently, we have $10\times 100=1,000$ ground-truthed frames with 7,770 word bounding boxes
for the dataset.
\par

\section{Lyric Word Detection and Tracking by Using Lyric Information\label{sec:track-method}}
In this section, we will introduce the methodology to detect and track lyric words in a lyric video.
The technical highlight of the methodology is to fully utilize the lyric information (i.e., the lyric word sequence of the song) for accurate tracking results. Note that it is assumed that the lyric word sequence is provided for each lyric video as metadata.

\begin{figure}[t!]
\centering
  \begin{minipage}[t]{0.8\linewidth}
    \centering
    \includegraphics[width=\textwidth]{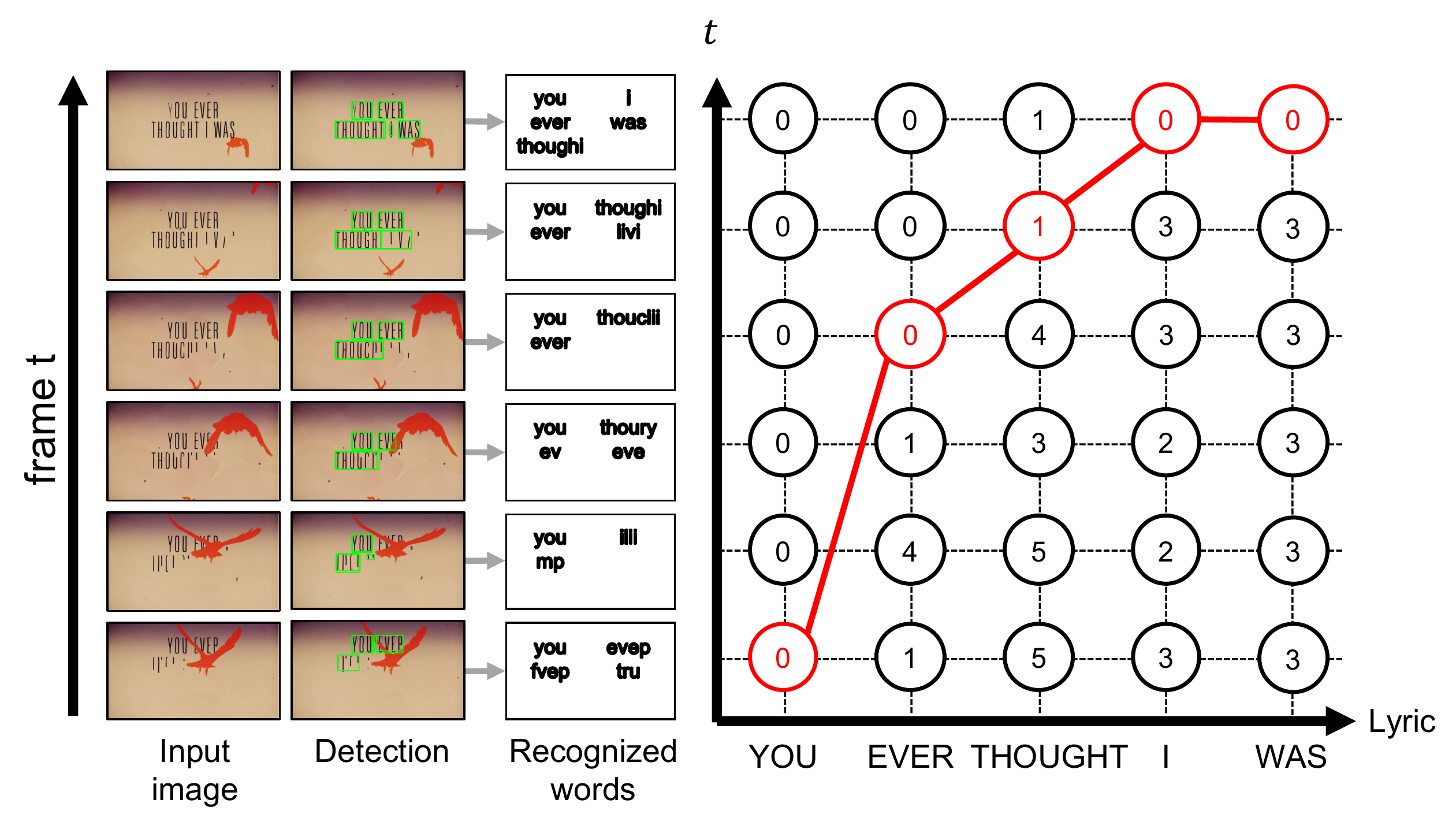}\\[-2mm]
    \subcaption{Detection, recognition, and lyric-frame matching.}
    \label{lyric_words_example1}
  \end{minipage}\par\bigskip
  \begin{minipage}[t]{0.4\linewidth}
    \centering
    \includegraphics[width=0.8\textwidth]{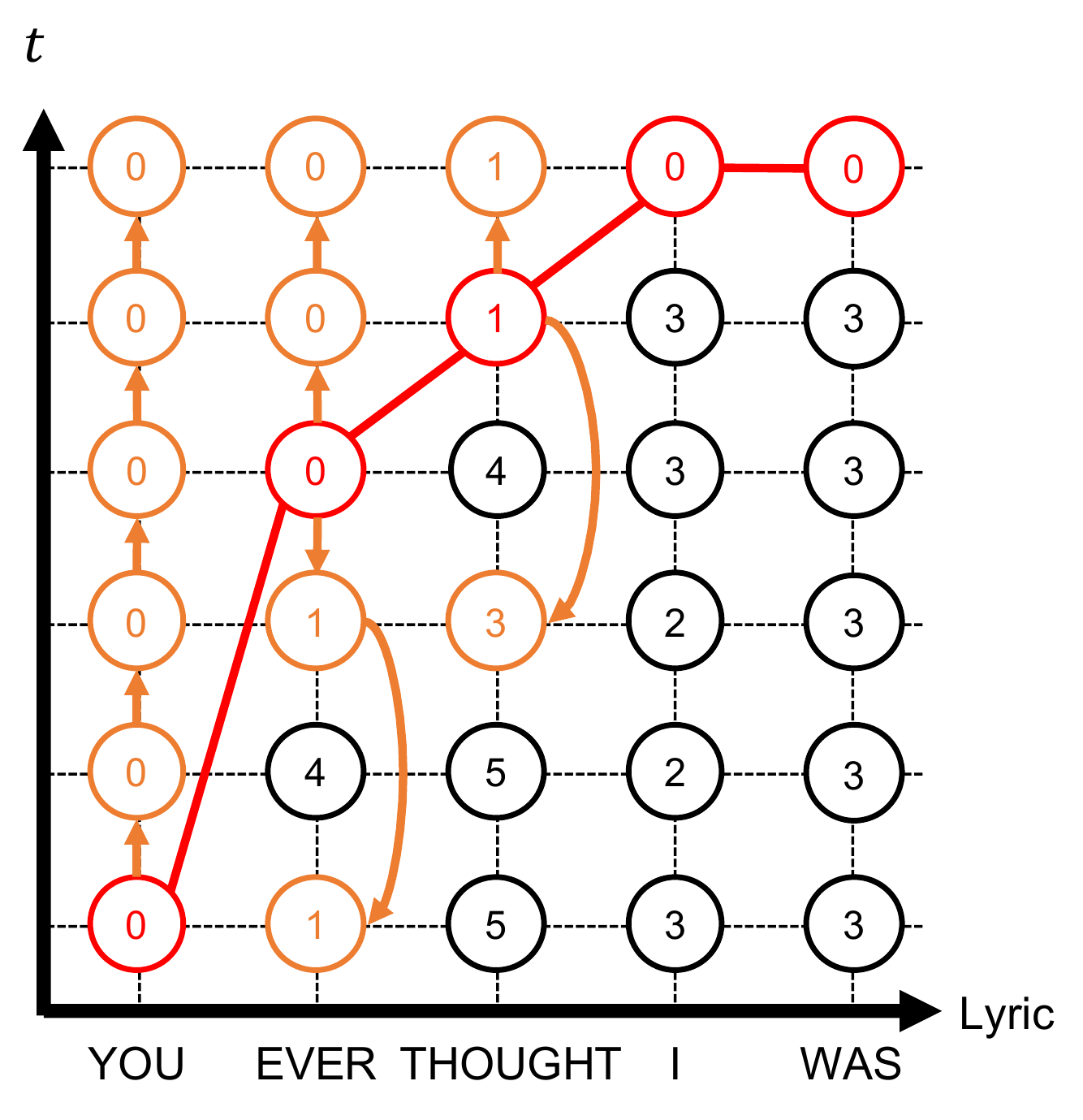}\\[-2mm]
    \subcaption{Tracking by neighbor search.}
    \label{lyric_words_example2}
  \end{minipage}
   \begin{minipage}[t]{0.4\linewidth}
    \centering
    \includegraphics[width=0.8\textwidth]{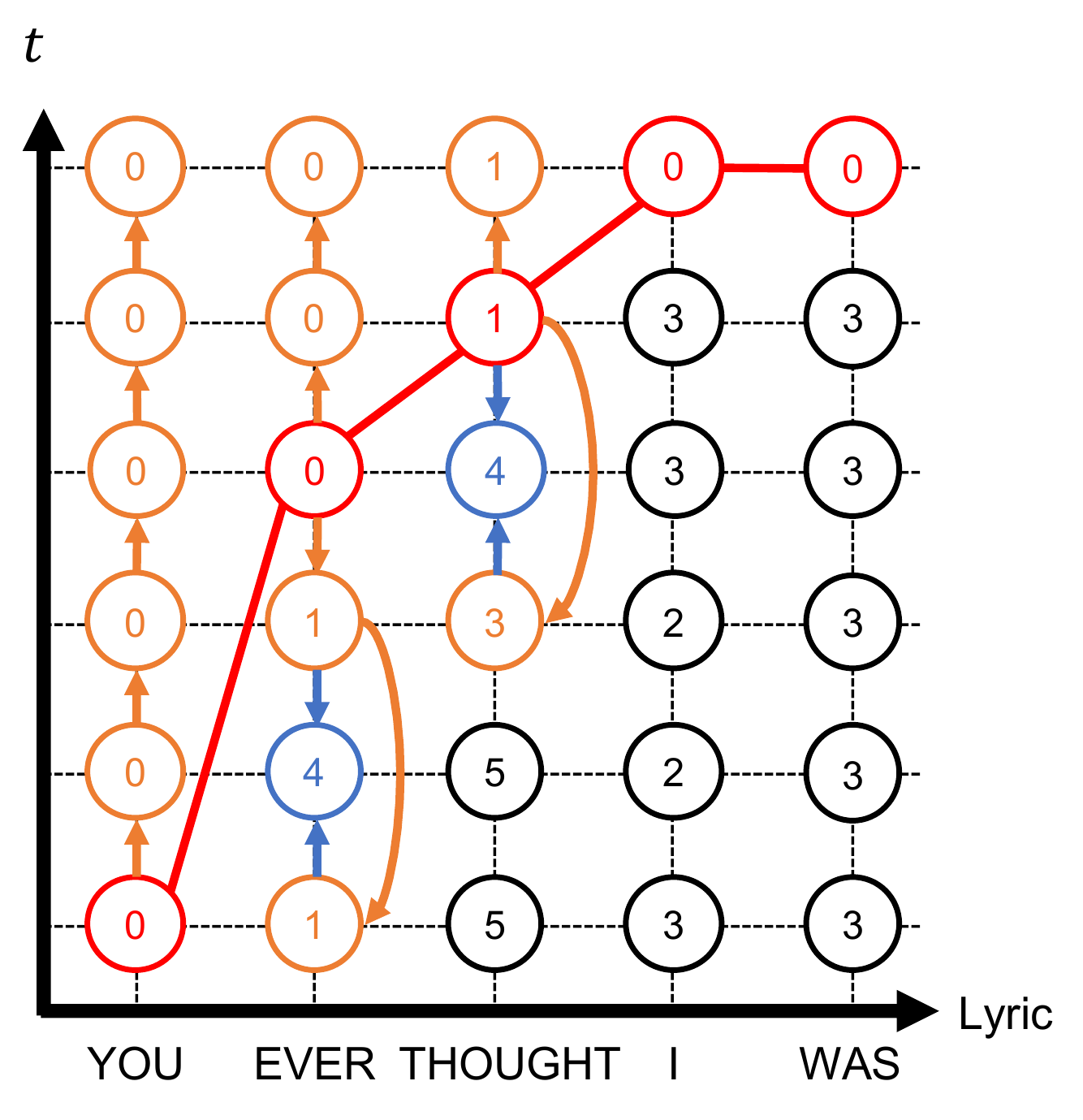}\\[-2mm]
    \subcaption{Interpolation.}
    \label{lyric_words_example2}
  \end{minipage}\par\bigskip
  \begin{minipage}[t]{0.8\linewidth}
    \centering
    \includegraphics[width=0.8\textwidth]{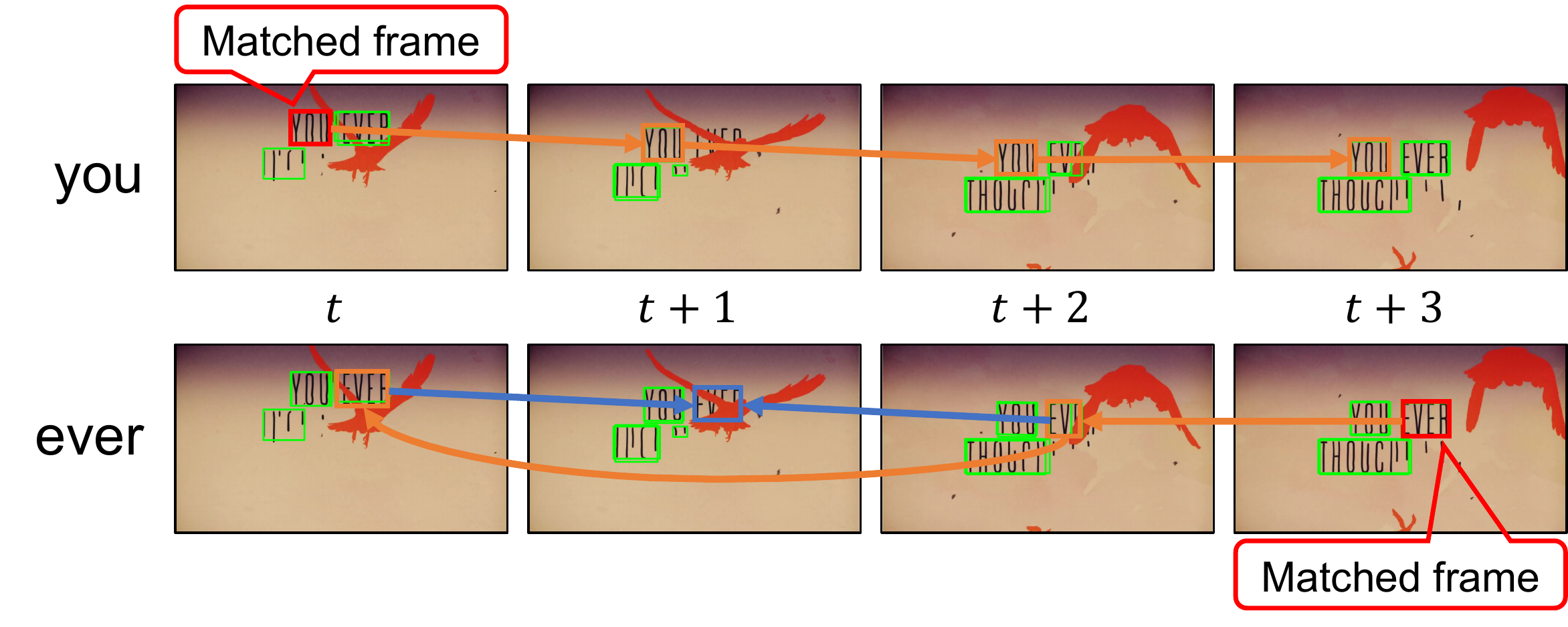}
    \subcaption{Tracking result of ``you'' and ``ever''.}
    \label{lyric_words_example2}
  \end{minipage}
  \caption{The detail of lyric word detection and tracking.}
  \label{fig:detail_tracking}
\end{figure}

\subsection{Lyric Word Candidate Detection\label{sec:detection}}
As the first step, lyric word candidates are detected by scene text detectors,
as shown in the left side of Fig.~\ref{fig:detail_tracking}~(a).
Specifically, we use two pretrained state-of-the-art detectors, PSENet~\cite{wang2019shape} and CRAFT~\cite{baek2019character}.
Then, each detected bounding box is fed to a state-of-the-art scene text recognizer; we used TPS-Resnet-BiLSTM-Attn, which achieved the best performance in ~\cite{baek2019wrong}.
If two bounding boxes from those two detectors
have 50\% overlap and the same recognition result, they are treated as the duplicated bounding boxes and thus one box is removed from the later process.

\subsection{Lyric-Frame Matching}
After the detection, lyric-frame matching is performed to find
the correspondence between the given lyric word sequence and the frame
sequence.
The red path on the right side of
Fig.~\ref{fig:detail_tracking}~(a) shows the matching path showing the optimal
correspondence.
Assume the video and its lyrics are comprised of $T$ frames and $K$ words,
respectively, and  $t\in [1,T]$  and $k\in [1,K]$ are their
indices, respectively. If the path goes through the node $(k,t)$,
the frame $t$ is determined as the most confident frame for the $k$th
lyric word. Note that the other frames where the same $k$th lyric word
appears will be determined later by the tracking process of Section~\ref{sec:tracking}.
\par
The path is determined by evaluating the distance $D(k,t)$ between the $k$th word and the frame $t$.
The circled number in Fig.~\ref{fig:detail_tracking}~(a)
represents $D(k,t)$. A smaller value of $D(k,t)$ means that the $k$th lyric word will be found at $t$ with a high probability.
The distance is calculated as $D(k,t)=\min_{b\in B_t} d(k,b)$, where
$B_t$ is the set of bounding boxes (i.e., words) detected at $t$ and
$d(k,b)$ is the edit distance function between the $k$th lyric word and the
$b$th detected word at $t$. If $D(k,t)=0$, the $k$th lyric word is correctly
detected at $t$ without misrecognition.\par
With the distance $\{D(k,t)|\forall k, \forall t\}$, the globally optimal lyric-frame matching is obtained efficiently by dynamic
programming (DP), as the red path of
Fig.~\ref{fig:detail_tracking}~(a). In the algorithm (so-called the DTW algorithm), the following DP recursion is
calculated at each $(k,t)$ from $(k,t)=(1,1)$ to $(K,T)$;
$$
    g(k,t) = D(k,t) + \min_{t-\Delta\leq t'< t} g(k-1,t'),
$$
where $g(k,t)$ becomes the minimum accumulated distance  from
$(1,1)$ to $(k,t)$. The parameter $\Delta$ specifies the maximum skipped frames
on the path. In the later experiment, we set $\Delta=1,000$ and it means that we
allow about 40-second skip for videos with 24 fps.
The computational complexity of the DTW algorithm is $O(\Delta TK)$.
\par
It should be emphasized that this lyric-frame matching process with the lyric
information is mandatory for lyric videos. For example, the word ``the'' will
appear in the lyric text many times; this means that there is a
large ambiguity of the spatio-temporal location of a certain ``the.'' We, therefore,
need to fully utilize the sequential nature of not only video frames but also lyric words
by the lyric-frame matching process for determining the
most confident frame for each lyric word.

\subsection{Tracking of Individual Lyric Words\label{sec:tracking}}
Although the $k$th lyric word is matched only to a single frame $t$ by the above lyric-frame
 matching step, the word will also appear in the neighboring frames of $t$. We,
 therefore, search those frames around the $t$th frame, as shown in
 Fig.~\ref{fig:detail_tracking}~(b). This search is simply done by evaluating
not only spatio-temporal similarity but also word similarity to the $k$th word, at the neighboring frames of $t$. If both similarities are larger than thresholds at $t'$, we assume the same $k$th word is also found at $t'$.\par
Finally, an interpolation process, shown in Fig.~\ref{fig:detail_tracking}~(c),
is performed for completing the spatio-temporal
 tracking process of each lyric word. By the above simple searching process,
the lyric word will be missed at a certain frame when a severe misrecognition
and/or occlusion occurs on the word at the frame.
In the example of Fig.~\ref{fig:detail_tracking}, the third lyric word
 ``THOUGHT'' is determined at the fifth frame by the lyric-frame matching
 process and then found at the third and sixth
 frames by the search. However, it was not found at the fourth frame due to
a severe misrecognition of the word.
If such a missed frame is found, the polynomial interpolation process
determines the location of the lyric word at the frame.
Fig.~\ref{fig:detail_tracking}~(d) shows the final result of the tracking
 process for two lyric words ``you'' and ``ever.''

\begin{figure}[t]
    \centering
    \includegraphics[width=1\textwidth]{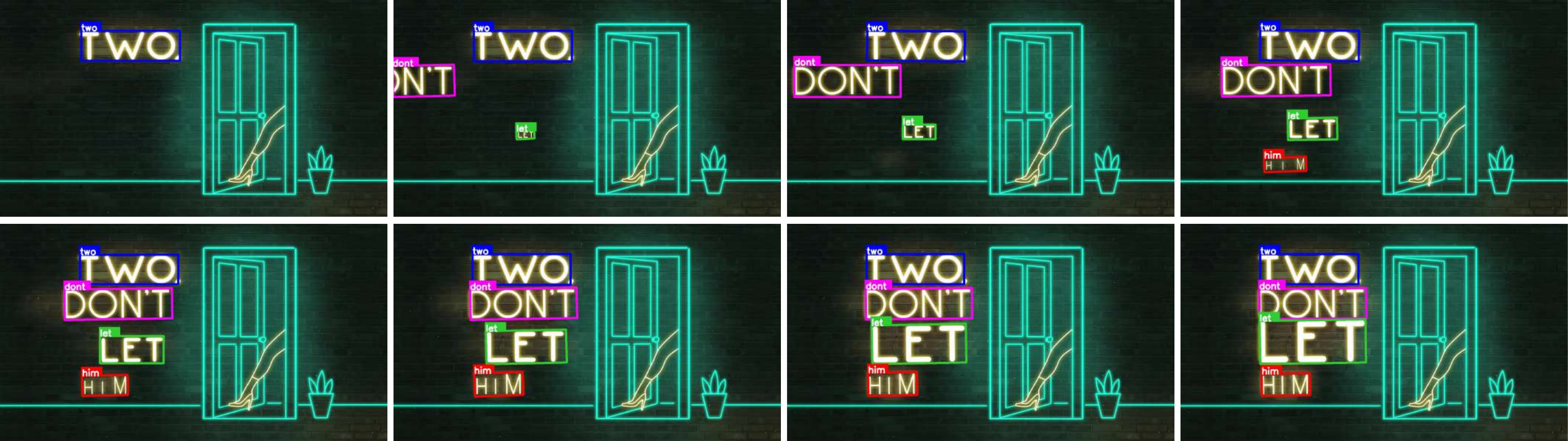}\\
    (a)~Motion by translation (``DON'T'').\\
    \medskip
    \includegraphics[width=1\textwidth]{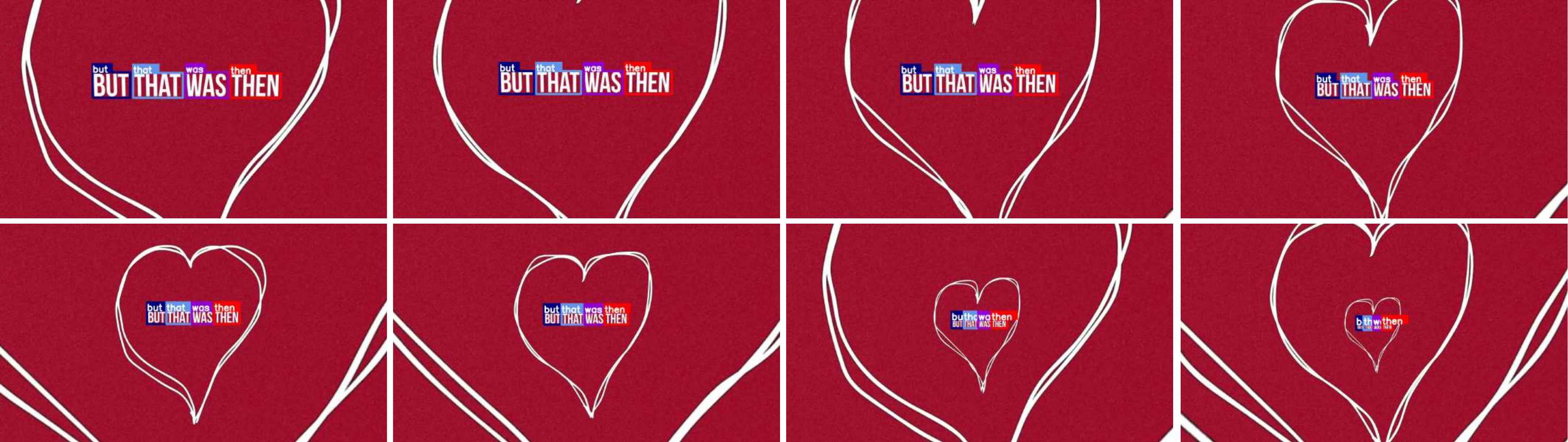}\\
    (b)~Motion by scaling.\\
    \medskip
    \includegraphics[width=1\textwidth]{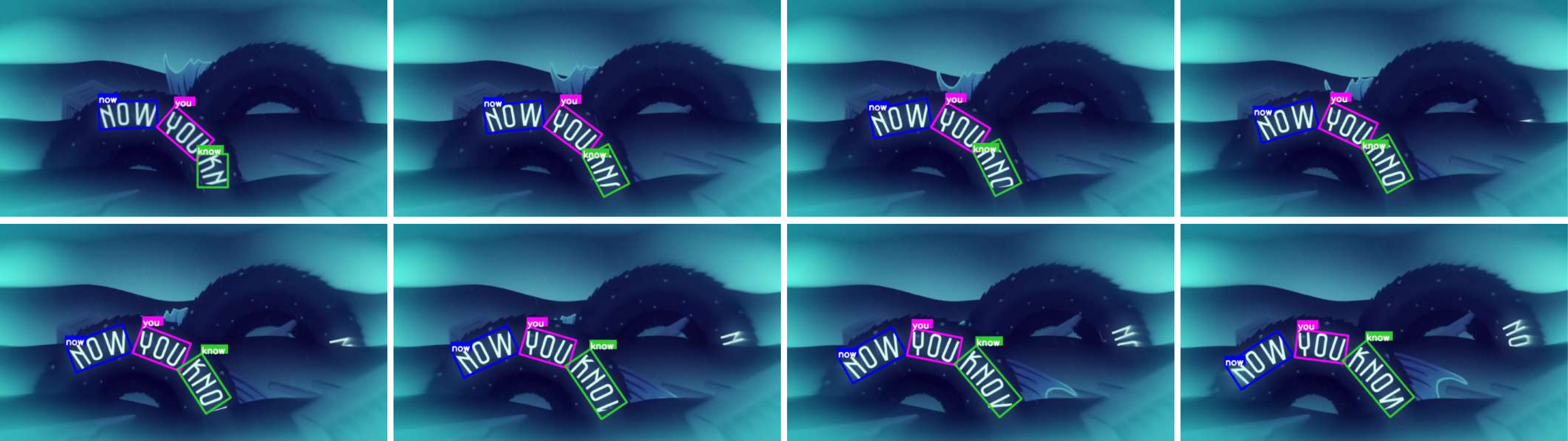}\\
    (c)~Motion by rotating.\\
  \caption{Successful results of lyric word detection and tracking under
 variable motion types. }
  \label{fig:successful_results}
\end{figure}

\begin{figure}[t!]
    \centering
    \includegraphics[width=1\textwidth]{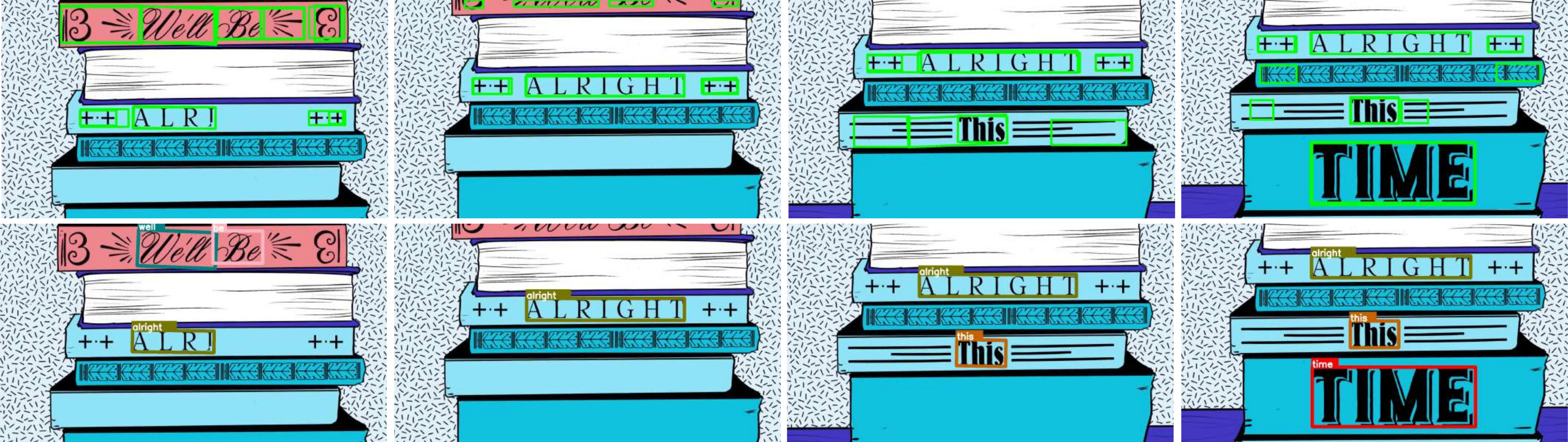}\\[-2mm]
    \caption{The effect of lyric information. Top: The initial word detection
 results shown as green boxes. Bottom: The final tracking result shown as blue boxes. The lyric words in those frames are ``well be alright this time''.}
    \label{fig:refinement_effect}
\bigskip
  \begin{minipage}[t]{1\linewidth}
    \centering
    \includegraphics[width=1\textwidth]{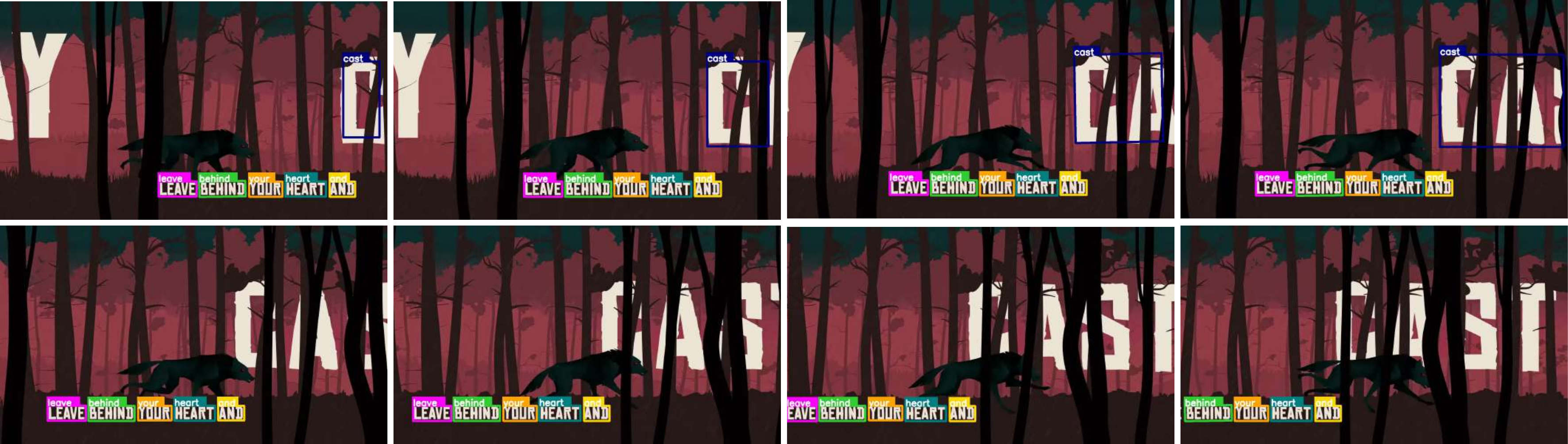}\\[-2mm]
    \subcaption{Heavy distortion by partial occlusion (``CAST'').}
    \label{failure_result1}
  \end{minipage}
  \begin{minipage}[t]{1\linewidth}
    \centering
    \includegraphics[width=1\textwidth]{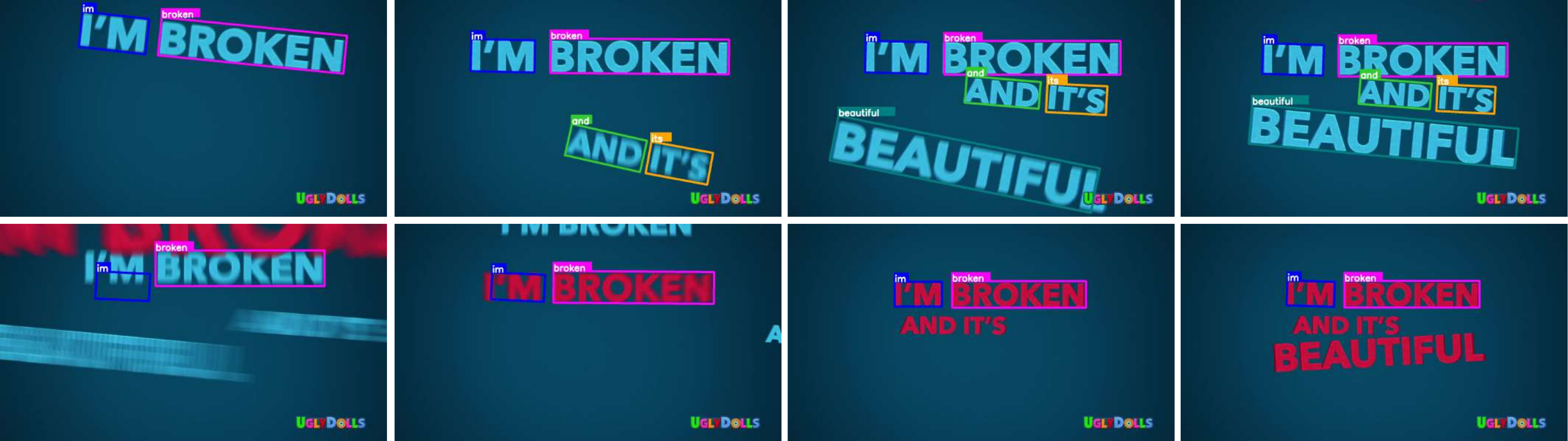}\\[-2mm]
    \subcaption{Tracking error due to multiple appearances of the same word (``BROKEN'').}
    \label{failure_result2}
  \end{minipage}
  \caption{Failure results of lyric word detection and tracking.}
  \label{fig:failure_results}
\end{figure}

\section{Experimental Results\label{sec:exp}}
By applying the proposed method to all frames of the collected 100 lyric videos (about 547,100 frames in total), we had the tracking results for all lyric words (about 33,800 words in total).
\subsection{Qualitative Evaluation}
 Fig.~\ref{fig:successful_results} shows several successful results of lyric word
detection and tracking. In~(a), the word ``DON'T'', which moves horizontally,
is tracked successfully. The words ``LET'' and ``HIM'' in (a) and all the words
in (b) are tracked correctly, although their size varies.
The words in (c) are also tracked successfully even under frame-by-frame rotation.\par

Fig.~\ref{fig:refinement_effect} confirms the effect of using lyric information at the lyric-frame matching process and
the following tracking process for accuracy improvement.
Since these frames have a complex background
(with character-like patterns), unnecessary bounding boxes are found by the
initial word detection step. In contrast, only the correct lyric words remain
after lyric-frame matching and tracking. \par

Fig.~\ref{fig:failure_results} shows typical failure cases. The failure of
(a) was caused by a heavy distortion on a word by an elaborated visual design of the video.
Specifically, the word ``CAST'' is always partially occluded
and never detected and recognized correctly over frames. Lyric videos frequently
show words which are difficult to read even by the state-of-the-art word
detectors and recognizers (and also by a human). The failure of (b) is caused by a
refrain of the same word in the lyrics. Especially in musical lyrics, it is very
frequent that the same word appears repeatedly within a short period. These
multiple appearances easily distract our tracking process.

\begin{table}[t]
  \centering
\caption{Performance of the lyric word detection and tracking. MA: Lyric-frame matching. TR: Tracking. IN: Interpolation. TP: \#True-positive. FP: \#False-positive. FN: \#False-negative. P: Precision (\%). R: Recall (\%). F: F-measure.}
 \smallskip
 \label{table:quantitative}
    \begin{tabular}{ccc||r|r|r|r|r|r}
    MA&TR&IN &\multicolumn{1}{|c|}{TP} &\multicolumn{1}{|c|}{FP} & \multicolumn{1}{|c|}{FN} & \multicolumn{1}{|c|}{P $=\frac{\mathrm{TP}}{\mathrm{TP}+\mathrm{FP}}$}& \multicolumn{1}{|c|}{R$=\frac{\mathrm{TP}}{\mathrm{TP}+\mathrm{FN}}$}& \multicolumn{1}{|c}{F$=\frac{2\mathrm{P}\mathrm{R}}{\mathrm{P}+\mathrm{R}}$} \\ \hline
     \checkmark & & & 72 & 12 & 7,698 & 85.71 & 0.93 & 0.0183 \\
     \checkmark & \checkmark&  & 5,513 & 462 & 2,257 & 92.27 & 70.95 & 0.8022 \\
     \checkmark & \checkmark & \checkmark  & 5,547 & 550 & 2,223 & 90.98 & 71.39 & 0.8000
    \end{tabular}
\end{table}
\subsection{Quantitative Evaluation\label{sec:quantitative}}
Table~\ref{table:quantitative} shows the quantitative evaluation result of the
performance of the lyric word detection and tracking, by using 1,000
ground-truthed frames. If a bounding box of a lyric word by the
proposed method and the ground-truth bounding box of the same lyric
word have IoU $>0.5$, the detected box is counted as a successful result.
If multiple detected boxes have an overlap with the same ground-truth box
with IoU $>0.5$, the detected box with the highest IoU is counted as a true-positive and the remaining are false-positives. \par

The evaluation result shows that the precision is
90.98\% and therefore the false positives are suppressed successfully by the proposed method with the lyric-frame matching step and the later tracking step. The interpolation step could increase TPs as expected, but, unfortunately, also increases FPs and slightly degrades the precision value.\par
The recall is about
71\% at the best setup\footnote{As explained before, only a single frame is determined for each lyric word by the lyric-frame matching. This results in very low recall (0.93\%).}. There are several reasons that degrade recall.
First, as shown in Fig.~\ref{fig:failure_results}~(a), the
elaborated visual design often disturbs the word detection and recognition
process and degrades the recall. In other words, we need to develop new
text detectors and recognizers that are robust to the visual disturbances
to the words in lyric videos as future work.\par
The second reason for the low recall is the errors at the lyric-frame matching
step. Since our lyric-frame matching allows the skip
of $\Delta=1,000$ frames (about 40 seconds) at maximum, to deal with frames
without lyrics (interlude). This sometimes causes a wrong skip of the frames
where lyric words actually appear and induce errors in the matching step.
Multiple appearances of the same word (such as ``I'') also induce matching errors.
\par
The third reason is the deviation from the official lyrics; even in the official
lyric videos, the lyrics shown in video frames are sometimes different from the
official lyrics due to, for example, the improvisation of the singer.
Since we believe the official lyrics for lyric-frame matching, such deviations
disturb the matching process and cause the degradation of recall. \par

\begin{figure}[t!]
\includegraphics[width=1\textwidth]{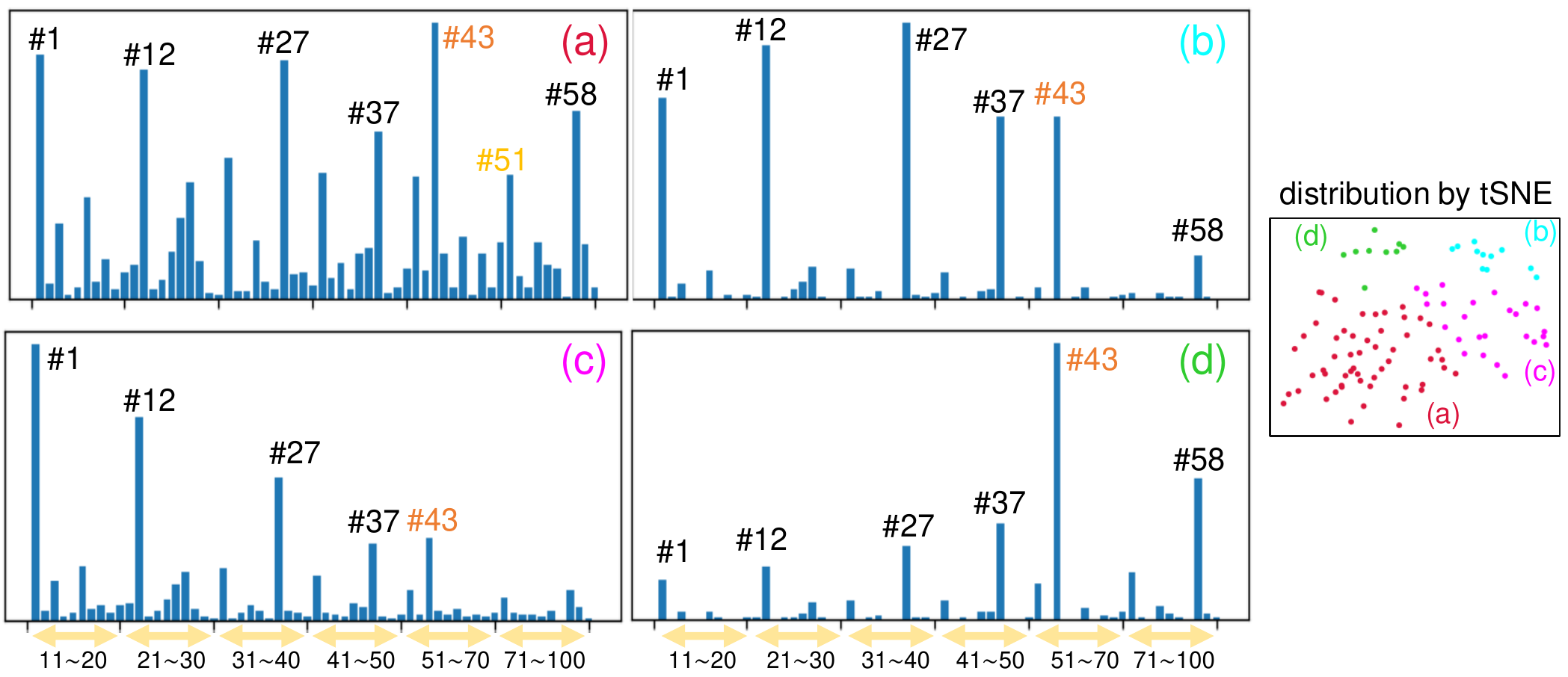}
\caption{The mean word motion histograms of four video clusters. Each of 60 bins corresponds to a representative word motion. ``11$\sim$20'' means the 10 representative motions for the trajectories with 11$\sim$20 frame length. At the right side, the distribution of the 100 histograms of the individual videos is visualized by tSNE.\label{fig:motion-hist}}
\end{figure}
\subsection{Text Motion Analysis by Clustering\label{sec:motion-analysis-result}}
Finally, we analyze the motion trajectory of lyric
words. As the result of the proposed lyric word detection and tracking process, we already
have the motion trajectory of individual lyric words. Trend analysis of those
trajectories will give a hint about how to design lyric videos. Note that we
consider that the trajectories subjected to this analysis are reliable enough
because the quantitative evaluation in Section~\ref{sec:quantitative} shows a
high precision. \par
As the motion trajectory analysis, we first determine representative motion trajectories via simple clustering.
Before the clustering, we divide them into six subsets by their length
($11\sim 20$ frames (4336 trajectories), $21\sim 30$ (4293), $31\sim 40$ (3797), $41\sim 50$ (3159), $51\sim 70$ (4057), $71\sim 100$ (2660)). Note that 6123 extremely short
($\leq 10$) and 2469 extremely long ($> 100$) trajectories are excluded from the analysis.
Then for each subset, we performed k-medoid clustering with DTW distances between trajectories.
We set $k=10$ and thus we have 60 representative motions.\par
Using those 60 representative motions, we create a histogram with 60 bins for each lyric video.
Specifically, like a bag-of-words, all word trajectories in the video are classified into one
of 60 representative motions and then we have a histogram whose $q$th bin represents the number of
the lyric words whose motion is the most similar to the $q$th representative motion. All histograms
are then normalized to have the same number of votes.\par
After this histogram-based representation of all videos, we perform another clustering (k-means)
in order to classify 100 videos into clusters with similar word motions. According to the Calinski-Harabasz
criterion, we have 4 clusters. Fig.~\ref{fig:motion-hist} shows the mean histogram of those clusters.
This means that our 100 lyric videos can be classified into four types and each of them contains
the lyric words whose motion trend is described by the corresponding histogram. Among 100 videos, 55
belong to (a), 11 to (b), 25 to (c), and 9 to (d).\par
In all the histograms, prominent peaks are found at \#1, \#12, \#27, \#37, and \#58 and all of those five representative
motions are static motion (with different frame lengths). This indicates that most lyric words are static.
With a closer inspection, we can find the ratios of the five representative motions are different in each histogram.
This indicates that the period of showing a word (without motion) is different by the tempo of the music. For example, (c) has many words displayed only for a very short (11$\sim$20) period.\par
Another prominent peak is found at \#43 and this represents a horizontal motion. It is interesting to note
that a peak at horizontal motions is only found at 51$\sim$70; this means that speed of the frequent
horizontal motions is almost constant regardless of lyric videos (and also regardless of their tempo). Another observation is that a video that belongs to (a) will contain wide motion varieties. For example, the \#51 representative motion is a circular motion.

\section{Conclusion and Future Work\label{sec:conclusion}}
To the authors' best knowledge, this paper is the first trial of analyzing lyric
videos as novel and dynamic documents. For this difficult analysis task, we
developed a novel technique, called lyric-frame matching, where the temporal
location, i.e., the frame, of each word in the lyrics is determined
automatically by dynamic programming-based optimization. We experimentally
showed that the combination of the
lyric-frame matching technique and several state-of-the-art word detectors and recognizers could detect lyric words with more than 90\% precision and 70\% recall on our original 100 official lyric video dataset.
Although the recall is lower than the precision, the current tracking performance is already reliable for analyzing the motion trajectories of lyric words for understanding the text motion design in
lyric videos. In fact, we could determine four typical motion patterns
of lyric videos for our 100 videos.
\par
Since this is the first trial of lyric video analysis, we have multiple future
works. First, word detection and recognition performance should be improved for
lyric videos. Since the distortions (including elaborated visual designs) on
lyric texts are often very different from those of scene texts. This means it is
still insufficient to apply state-of-the-art scene text detectors and
recognizers to this task.  The second and more important future work is to
analyze word motion trajectories in lyric videos. Since the moving words in
lyric videos are a new target of document analysis, we need to develop many
analysis schemes. Word motion analysis for each music genre is straightforward.
A more important analysis scheme is the correlation analysis between word motion
and music signals, i.e., sounds. We often observe that word motions are strongly
affected by the beat and mood of the music. Our final goal is to
generate lyric videos automatically or semi-automatically from lyric information
and music signals, based on those technologies and analyses.
For example, our motion trajectory analysis could contribute to enabling data-driven motion styles on a tool for creating lyric videos~\cite{KaNG2015a}.
\section*{Acknowledgment}
This work was supported by JSPS KAKENHI Grant Number JP17H06100.

\section*{Videos shown in the figures}
In the figures of this paper, the following video frames are shown. For URL, the common prefix part  ``{\tt https://www.youtube.com/watch?v=}'' is omitted in the list.
Note that the URL list of all 100 videos and their annotation data can be found at {\tt https://github.com/uchidalab/Lyric-Video}.
\par
\begin{itemize}
\item Fig.~\ref{fig:lyric_video_examples}: (a) MK，17, {\tt NoBAfjvhj7o}; (b) Demi Lovato, Really Don't Care,\\{\tt EOEeN9NmyU8}; (c) Alok, Felix Jaehn \& The Vamps，All The Lies, {\tt oc218bqEbAA}; (d) Taylor Swift, Look What You Made Me Do, {\tt 3K0RzZGpyds};\par
\item Fig.~\ref{fig:overall}: Robin Schulz, Sugar, {\tt jPW5A\_JyXCY}.\par
\item Fig.~\ref{fig:lyric_words_example}: (a) Marshmello x Kane Brown, One Thing Right, {\tt O6RyKbcpBfw}; (b) Harris J, I Promise, {\tt PxN6X6FevFw}; (c) Rita Ora, Your Song, {\tt i95Nlb7kiPo}; (d) Green Day, Too Dumb to Die, {\tt qh7QJ\_jLam0};\par
\item Fig.~\ref{fig:detail_tracking}: Freya Ridings, Castles, {\tt pL32uHAiHgU}.\par
\item Fig.~\ref{fig:successful_results}: (a) Dua Lipa, New Rules, {\tt AyWsHs5QdiY}; (b) Anne-Marie, Then, {\tt x9OJpU7O\_cU}; (c) Imagine Dragons, Bad Liar, {\tt uEDhGX-UTeI};\par
\item Fig.~\ref{fig:refinement_effect}: Ed Sheeran, Perfect, {\tt iKzRIweSBLA};\par
\item Fig.~\ref{fig:failure_results}: (a) Imagine Dragons, Natural, {\tt V5M2WZiAy6k}; (b) Kelly Clarkson, Broken \& Beautiful, {\tt 6l8gyacUq4w};\par
\end{itemize}
\bibliography{hoge}

\begin{thebibliography}{10}

\bibitem{yin2016text}
X.~C. Yin, Z.~Y. Zuo, S.~Tian, and C.~L. Liu.
\newblock Text detection, tracking and recognition in video: a comprehensive
  survey.
\newblock {\em IEEE TIP}, 2016.

\bibitem{soccer}
M~Bertini, A.~Del~Bimbo, and W.~Nunziati.
\newblock Automatic detection of player's identity in soccer videos using faces
  and text cues.
\newblock In {\em ACM Multimedia}, 2006.

\bibitem{ijdar}
X.~Liu, G.~Meng, and C.~Pan.
\newblock Scene text detection and recognition with advances in deep learning:
  a survey.
\newblock {\em IJDAR}, 2017.

\bibitem{zhao2010text}
X.~Zhao, K.~H. Lin, Y.~Fu, et~al.
\newblock Text from corners: a novel approach to detect text and caption in
  videos.
\newblock {\em IEEE TIP}, 2010.

\bibitem{zhong2000automatic}
Y.~Zhong, H.~Zhang, and A.~K. Jain.
\newblock Automatic caption localization in compressed video.
\newblock {\em IEEE TPAMI}, 2000.

\bibitem{yang2012caption}
Z.~Yang and P.~Shi.
\newblock Caption detection and text recognition in news video.
\newblock In {\em Int. Congress Image Sig. Proc.}, 2012.

\bibitem{lu2014video}
T.~Lu, S.~Palaiahnakote, C.~L. Tan, and W.~Liu.
\newblock Video caption detection.
\newblock In {\em Video Text Detection}. Springer, 2014.

\bibitem{yang2014framework}
H.~Yang, B.~Quehl, and H.~Sack.
\newblock A framework for improved video text detection and recognition.
\newblock {\em Multimedia Tools and Applications}, 2014.

\bibitem{zhong2016recognition}
D.~Zhong, P.~Shi, D.~Pan, and Y.~Sha.
\newblock The recognition of {Chinese} caption text in news video using
  convolutional neural network.
\newblock In {\em IEEE IMCEC}, 2016.

\bibitem{zedan2016caption}
I.~A. Zedan, K.~M. Elsayed, and E.~Emary.
\newblock Caption detection, localization and type recognition in arabic news
  video.
\newblock In {\em INFOS}, 2016.

\bibitem{chen2018video}
L.~H. Chen and C.~W. Su.
\newblock Video caption extraction using spatio-temporal slices.
\newblock {\em Int. J. Image and Graphics}, 2018.

\bibitem{xu2018end}
Y.~Xu, et~al.
\newblock End-to-end subtitle detection and recognition for videos in {East
  Asian languages via CNN} ensemble.
\newblock {\em Signal Processing: Image Communication}, 2018.

\bibitem{lu2018novel}
W.~Lu, H.~Sun, J.~Chu, X.~Huang, and J.~Yu.
\newblock A novel approach for video text detection and recognition based on a
  corner response feature map and transferred deep convolutional neural
  network.
\newblock {\em IEEE Access}, 2018.

\bibitem{qian2007text}
X.~Qian, G.~Liu, H.~Wang, and R.~Su.
\newblock Text detection, localization, and tracking in compressed video.
\newblock {\em Signal Processing: Image Communication}, 2007.

\bibitem{Nguyen2014}
P.~X. Nguyen, K.~Wang, and S.~Belongie.
\newblock {Video text detection and recognition: Dataset and benchmark}.
\newblock In {\em WACV}, 2014.

\bibitem{gomez2014mser}
L.~G{\'o}mez and D.~Karatzas.
\newblock {MSER}-based real-time text detection and tracking.
\newblock In {\em ICPR}, 2014.

\bibitem{wu2015new}
L.~Wu, P.~Shivakumara, T.~Lu, and C.~L. Tan.
\newblock A new technique for multi-oriented scene text line detection and
  tracking in video.
\newblock {\em IEEE Trans. Multimedia}, 2015.

\bibitem{tian2017unified}
S.~Tian, XC. Yin, Y.~Su, and HW. Hao.
\newblock A unified framework for tracking based text detection and recognition
  from web videos.
\newblock {\em IEEE TPAMI}, 2017.

\bibitem{yang2017tracking}
C.~Yang, X.~C. Yin, W.~Y. Pei, et~al.
\newblock Tracking based multi-orientation scene text detection: A unified
  framework with dynamic programming.
\newblock {\em IEEE TIP}, 2017.

\bibitem{pei2018scene}
W.~Y. Pei, et~al.
\newblock Scene video text tracking with graph matching.
\newblock {\em IEEE Access}, 2018.

\bibitem{wang2018robust}
Y.~Wang, L.~Wang, and F.~Su.
\newblock A robust approach for scene text detection and tracking in video.
\newblock In {\em Pacific Rim Conf. Multimedia}, 2018.

\bibitem{wang2019video}
Y.~Wang, L.~Wang, F.~Su, and J.~Shi.
\newblock Video text detection with fully convolutional network and tracking.
\newblock In {\em ICME}, 2019.

\bibitem{movingmnist}
N.~Srivastava, E.~Mansimov, and R.~Salakhutdinov.
\newblock Unsupervised learning of video representations using {LSTMs}.
\newblock In {\em ICML}, 2015.

\bibitem{wang2019shape}
W.~Wang, E.~Xie, X.~Li, et~al.
\newblock Shape robust text detection with progressive scale expansion network.
\newblock In {\em CVPR}, 2019.

\bibitem{baek2019character}
Y.~Baek, B.~Lee, D.~Han, S.~Yun, and H.~Lee.
\newblock Character region awareness for text detection.
\newblock In {\em CVPR}, 2019.

\bibitem{baek2019wrong}
J.~Baek, G.~Kim, J.~Lee, et~al.
\newblock What is wrong with scene text recognition model comparisons? dataset
  and model analysis.
\newblock In {\em ICCV}, 2019.

\bibitem{KaNG2015a}
J.~Kato, T.~Nakano, and M.~Goto.
\newblock {TextAlive}: Integrated design environment for kinetic typography.
\newblock In {\em ACM CHI}, 2015.

\end{thebibliography}
\bibliographystyle{junsrt}

\end{document}